\documentclass[]{fairmeta}

\title{\vspace{-1pt}ReconDrive: Fast Feed-Forward 4D Gaussian Splatting for Autonomous Driving Scene Reconstruction}

\author{Haibao Yu$^{2,1,*,\ddagger}$
Kuntao Xiao$^{1,*}$, Jiahang Wang$^{1}$, Ruiyang Hao$^{3,1}$, Yuxin Huang$^{4,1}$, Guoran Hu$^{5,1}$, Haifang Qin$^{1}$, Bowen Jing$^{1}$, Yuntian Bo$^{1}$, Ping Luo$^{2}$ 
\vspace{-3ex}
}

\affiliation[]{
 \\
\text{$^{1}$Tuojing Intelligence} \\
\text{$^{2}$The University of Hong Kong, $^{3}$King's College London} \\
\text{$^{4}$The University of Sydney, $^{5}$Mohamed bin Zayed University of Artificial Intelligence} 
}

\contribution[*]{equal contribution}
\contribution[\ddagger]{corresponding author}

\usepackage{amsmath,amsfonts,bm}
\usepackage{xcolor}

\def\eqref#1{equation~\ref{#1}}

\def\1{\bm{1}}

\DeclareMathAlphabet{\mathsfit}{\encodingdefault}{\sfdefault}{m}{sl}
\SetMathAlphabet{\mathsfit}{bold}{\encodingdefault}{\sfdefault}{bx}{n}

\usepackage{graphicx}
\usepackage[table]{xcolor}
\usepackage{colortbl}

\usepackage{booktabs}
\usepackage{tabularx}
\usepackage{array}
\usepackage{multirow}
\usepackage{diagbox}
\usepackage{hhline}
\usepackage{longtable}
\usepackage{makecell}
\usepackage{siunitx}
\usepackage{adjustbox}
\usepackage{wrapfig}
\usepackage{caption}   
\newcolumntype{C}{>{\centering\arraybackslash}X}

\usepackage{amsmath,amsfonts,amssymb}
\usepackage{bm}
\usepackage{nicefrac}

\usepackage{enumitem}
\setlist[itemize]{leftmargin=*}

\usepackage{caption}
\crefname{figure}{Fig.}{Figs.}
\crefname{table}{Tab.}{Tabs.}

\usepackage{xspace}
\usepackage{calc}
\usepackage{etoolbox}
\usepackage{pifont}
\usepackage{fancyvrb}

\usepackage{titletoc}

\titlecontents{section}
[1.5em] %
{\addvspace{-0.5pt}} %
{\bfseries\contentslabel{2.3em}} %
{\hspace*{-2.3em}\bfseries} %
{\bfseries\titlerule*[.5pc]{.}\contentspage} %
\titlecontents{subsection}
[3.8em] %
{\addvspace{-2.2pt}} %
{\contentslabel{2.3em}}
{\hspace*{-2.3em}}
{\titlerule*[.5pc]{.}\contentspage}

\abstract{
	High-fidelity visual reconstruction and novel-view synthesis are essential for realistic closed-loop evaluation in autonomous driving. While 4D Gaussian Splatting (4DGS) offers a promising balance of accuracy and efficiency, existing per-scene optimization methods require costly iterative refinement, rendering them unscalable for extensive urban environments. Conversely, current feed-forward approaches often suffer from degraded photometric quality. To address these limitations, we propose ReconDrive, a feed-forward framework that leverages and extends the 3D foundation model VGGT for rapid, high-fidelity 4DGS generation.
    Our architecture introduces two core adaptations to tailor the foundation model to dynamic driving scenes: (1) Hybrid Gaussian Prediction Heads, which decouple the regression of spatial coordinates and appearance attributes to overcome the photometric deficiencies inherent in generalized foundation features; and (2) a Static-Dynamic 4D Composition strategy that explicitly captures temporal motion via velocity modeling to represent complex dynamic environments.
    Benchmarked on nuScenes across reconstruction, synthesis, and 3D perception tasks, ReconDrive outperforms all existing feed-forward baselines and surpasses per-scene optimization methods in eight out of nine evaluation metrics. By delivering superior quality with substantially fewer resources, our method provides a highly scalable and efficient solution for large-scale autonomous driving scene reconstruction and simulation.
	\vspace{-10pt}
}
\metadata[Code]{\url{https://github.com/TuojingAI/ReconDrive}}

\definecolor{lightgray}{rgb}{0.95, 0.95, 0.95}

\definecolor{baselinecolor}{gray}{.9}

\begin{document}
\maketitle
\section{Introduction}
\label{sec:intro}


Closed-loop evaluation is increasingly vital for end-to-end autonomous driving, requiring the ego-vehicle to receive sensor observations that faithfully correspond to its actions within a simulated environment. Visual scene reconstruction and novel-view synthesis from real-world multi-view imagery provide the bedrock for such simulations. Existing approaches generally fall into three categories: asset importation into game engines like CARLA~\cite{dosovitskiy2017carla}, implicit representation via NeRF~\cite{mildenhall2021nerf} or video diffusion, and explicit geometric modeling. Among these, 4D Gaussian Splatting (4DGS)~\cite{wu20244d}, an extension of 3DGS~\cite{kerbl20233d} with a temporal dimension, effectively balances geometric accuracy, photometric fidelity, and real-time rendering, making it ideal for interactive driving simulations.

However, traditional 4DGS methods, such as Street Gaussian~\cite{yan2024street}, rely on per-scene optimization. These approaches iteratively optimize Gaussian kernels to minimize rendering loss, often requiring LiDAR priors for initialization. This non-data-driven paradigm fails to leverage shared structural knowledge across scenes and incurs substantial computational overhead for every new environment, limiting scalability. In contrast, feed-forward models like VGGT~\cite{wang2025vggt} demonstrate the potential of 3D foundation models for efficient reconstruction by directly producing explicit geometry. Yet, extending VGGT to 4D Gaussian reconstruction for autonomous driving presents three primary challenges. First is photometric deficiency, as VGGT features lack the fine-grained detail necessary to regress high-fidelity appearance attributes. Second is temporal staticity, where static backbones cannot represent the dynamic motion of traffic participants. Third is domain and calibration mismatch, where data gaps lead to 3D geometry prediction errors and a failure to utilize the pre-calibrated sensor intrinsics and extrinsics inherent in driving datasets.

\begin{figure*}[t]
    \setlength{\belowcaptionskip}{0pt}
	\centering
	\includegraphics[width=1.0\textwidth]{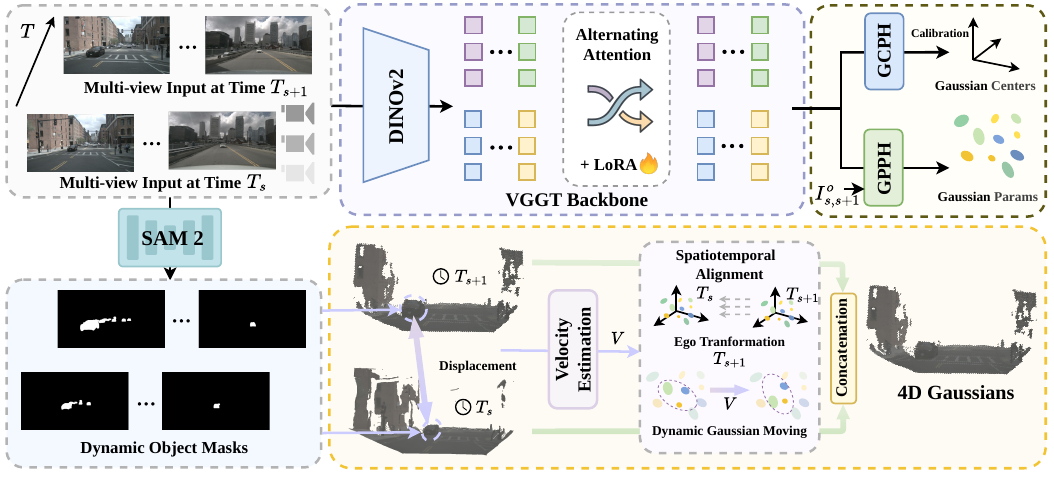}
  \vspace{-9pt}
	\caption{\textbf{ReconDrive Inference Framework.} ReconDrive is a powerful feed-forward 4D Gaussian splatting generation framework tailored for urban scene reconstruction and novel-view synthesis. In this framework, we select two context frames from each segment of an urban scene as input, and adopt a static-dynamic composition strategy to represent 4D Gaussians. To further enhance geometric precision, we design a dedicated Gaussian prediction head, consisting of a Gaussian Parameter Prediction Head (GPPH) and a Gaussian Center Prediction Head (GCPH), which enables the generation of Gaussians with more accurate geometric attributes.}
 \label{fig:ReconDrive-framework}
     \vspace{-8pt}
\end{figure*}

To address these limitations, we propose ReconDrive, a feed-forward framework that reconstructs 4D Gaussian Splatting representations from vision-only inputs. ReconDrive utilizes a pre-trained VGGT-style backbone to directly regress spatio-temporal Gaussian parameters, enabling efficient reconstruction at scale without per-scene optimization. The inference pipeline is built upon three design pillars. First, we employ hybrid Gaussian prediction heads; we concatenate raw images with dense pixel-wise features to provide the photometric cues required for regressing appearance attributes like opacity and spherical harmonics, while incorporating camera calibration directly into the centers head for precise spatial positioning. Second, we utilize a static-dynamic 4D composition, leveraging the SAM2~\cite{ravi2024sam} foundation model to extract object masks and assign temporal velocities to dynamic Gaussians through pixel-wise correspondence. Third, we implement segment-wise temporal fusion to handle view sparsity over long driving sequences by partitioning the scene into segments and fusing localized Gaussian clusters into a unified 4D representation.
The overall inference framework is illustrated in Fig.~\ref{fig:ReconDrive-framework}.
During training, we utilize frozen VGGT~\cite{wang2025vggt} weights and apply LoRA~\cite{hu2022lora} for parameter-efficient fine-tuning on urban driving data. We introduce a novel-frame projection loss to facilitate stable geometry reconstruction against unseen temporal views.

To ensure a fair comparison, we establish a comprehensive benchmark on the nuScenes~\cite{caesar2020nuscenes} dataset, reproducing five representative optimization and feed-forward methods under a unified protocol. We evaluate performance across scene reconstruction, novel-view synthesis, and downstream tasks such as 3D object detection and tracking. Within this benchmark, ReconDrive significantly outperforms existing feed-forward approaches across all metrics and surpasses per-scene optimization methods in eight out of nine categories. Our results demonstrate that feed-forward approaches can achieve competitive and often superior novel-view synthesis performance while requiring substantially fewer resources for Gaussian generation. This highlights the promising potential of feed-forward paradigms for large-scale autonomous driving reconstruction.

Our contributions can be summarized as follows.
\begin{itemize}[leftmargin=9pt]
\item We introduce ReconDrive, a feed-forward framework that can directly generate 4D Gaussian splatting without per-scene optimization, to reconstruct urban scenes and enable novel-view synthesis across time.
\item We introduce hybrid Gaussian prediction heads, a static-dynamic composition strategy, and a segment-wise temporal fusion representation. Combined with specialized training, these components enable 3D foundation models in dynamic urban environments.
\item We build autonomous driving scene reconstruction benchmark by reproducing typical baselines on the nuScenes~\cite{caesar2020nuscenes}. Our ReconDrive achieves state-of-the-art performance among feed-forward methods and competitive results against optimization-based baselines across multiple evaluation protocols.
\end{itemize}
\section{Related Work}
\label{sec:related-work}

\paragraph{Feed-Forward 3D and 4D Reconstruction.}
Feed-forward approaches have demonstrated significant potential in directly processing multi-view images to infer 3D geometric attributes. Foundational architectures such as DUSt3R~\cite{wang2024dust3r} and VGGT~\cite{wang2025vggt} bypass traditional optimization by directly predicting 3D point clouds and depth maps. To further enhance efficiency, FastVGGT~\cite{shen2025fastvggt} introduces region-based random sampling, which markedly reduces inference latency, particularly for large-scale inputs. Concurrently, increasingly powerful models like Concurrently, models like Depth Anything V3~\cite{lin2025depth} and SAM3D~\cite{chen2025sam} have advanced generalizable 3D scene understanding. 
Beyond point clouds, MVSplat~\cite{chen2024mvsplat} and AnySplat~\cite{jiang2025anysplat} extend this paradigm to Gaussian Splatting, though primarily for static environments.
Recent works have begun to address temporal dynamics within this paradigm. StreamVGGT~\cite{zhuo2025streaming} enables streaming reconstruction by maintaining a memory cache of previously computed tokens and leveraging temporal causal attention to accelerate inference over time. In the driving domain, DriveVGGT~\cite{jia2025drivevggt} utilizes the VGGT backbone within a scale-aware framework to output depth and point clouds specifically for autonomous driving data.
Our work shifts the focus from static geometric design to enhancing dynamic scene representation and novel-view fidelity. By adapting the VGGT formulation specifically for the complexities of autonomous driving, we achieve significant gains in visual quality and temporal consistency through improved feature utilization.

\paragraph{Gaussian Splatting for Autonomous Driving Scene Reconstruction.}
Existing methods primarily rely on per-scene optimization, iteratively refining Gaussian kernels to minimize rendering loss. AutoSplat~\cite{khan2025autosplat} achieves high-fidelity reconstruction by imposing geometric constraints on varying regions, while PVG~\cite{chen2023periodic} introduces periodic vibration Gaussians for dynamic urban environments. To handle the complexity of traffic, StreetGaussians~\cite{yan2024street} decomposes scenes into static backgrounds and moving vehicles, and DrivingGaussian~\cite{zhou2024drivinggaussian} introduces a dynamic Gaussian graph for moving objects. Similarly, S3Gaussian~\cite{huang2024textit} utilizes a spatial-temporal field network to model 4D dynamics. Despite their high quality, these optimization-based methods require hours of computation per scene, making them unscalable for large-scale autonomous driving datasets.
In contrast, the feed-forward paradigm generalizes to diverse scenes without requiring per-scene optimization. Drivingforward~\cite{tian2025drivingforward} reconstructs adjacent frames through interpolation, while STORM~\cite{yang2024storm} employs a transformer to generate per-frame Gaussians and scene flow via learnable motion tokens. However, these methods do not leverage 3D foundation models, which limits their representative capacity. More recently, WorldSplat~\cite{zhu2025worldsplat} introduced a 4D-aware latent diffusion model for pixel-aligned Gaussian production, and DGGT~\cite{chen2025dggt} proposed diffusion-based refinement to mitigate motion artifacts. Nevertheless, the integration of diffusion models significantly increases inference latency.
Unlike these approaches, ReconDrive leverages a 3D foundation model to enable scalable 4D Gaussian prediction with the well-designed architecture. By incorporating explicit motion and temporal consistency, we achieve efficient urban scene reconstruction without the need for iterative optimization or high-latency post-refinement.
\section{Method}
\label{sec:method}

We introduce ReconDrive, a feed-forward framework to process the urban scenes and generate the 4D gaussian splatting in one pass. We start by introducing the problem in Sec.~\ref{sec:method-problem}, followed by our model design in Sec.~\ref{sec:method-gs} and 4D gaussian splatting in Sec.~\ref{sec:method-4dgs}, and finally the training in Sec.~\ref{sec:method-training}.

\subsection{Problem Definition and Overall Inference}
\label{sec:method-problem}
\paragraph{Problem Formulation.} We formalize visual scene reconstruction and novel-view synthesis as the recovery of a continuous 4D representation from discrete observations. The input is an urban scene sequence $D = \{(I_t^o, \mathcal{C}^o, \mathcal{E}_t) \mid t \in [0, T], o=1,\dots,O\}$, where $I_t^o \in \mathbb{R}^{H \times W \times 3}$ is the RGB image captured at time $t$ by the $o$-th camera. Each camera is characterized by a static calibration matrix $\mathcal{C}^o$, which encapsulates both intrinsic parameters and the fixed extrinsic transformation relative to the ego-vehicle coordinate system. The temporal movement of the scene is captured by the time-varying ego-vehicle pose $\mathcal{E}_t \in \mathbb{SE}(3)$, which maps the vehicle's local coordinate frame to a global world frame. The objective is to utilize a feed-forward model to map $D$ to a 4D Gaussian Splatting representation that decouples static geometry from dynamic traffic participants. For any arbitrary timestamp $t' \in [0, T]$ and a novel viewpoint $\mathcal{E}_{t'}^{\text{novel}}$, the model retrieves and projects the corresponding temporal Gaussian kernels to render high-fidelity observations.

\paragraph{Challenges.} Developing a feed-forward 4DGS framework for autonomous driving entails three primary hurdles: 
1) Scalability \textit{vs.} Efficiency: Balancing extensive spatial coverage of large-scale urban environments with the requirements for real-time rendering.
2) Spatio-Temporal Modeling: Reconstructing dynamic scenes where 4D Gaussians must explicitly decouple and model motion, moving beyond the static assumptions of traditional multi-view foundation models.
3) Geometric-Photometric Alignment: Ensuring that regressed Gaussian attributes maintain both accurate 3D geometry for spatial consistency and high photometric fidelity for realistic sensor simulation.
In this work, we address these requirements by extending the capabilities of existing 3D foundation models. We propose a specialized architecture and training strategy that leverages the geometric priors of the foundation model while adapting them to the high-dynamic and calibrated nature of driving scenes.

\paragraph{Inference Pipeline Overview.} To maintain computational tractability over long-duration sequences, we adopt a segment-wise representation. The scene is partitioned into temporal segments $\{[T_{s}, T_{s+1}]\}$, with 4D Gaussians generated for each segment independently. This approach maximizes environmental coverage while limiting the active Gaussian count during rendering to ensure real-time performance. Within each segment, a 4D Gaussian kernel $\mathcal{G}_i(t)$ is defined by a center-moving paradigm:\begin{equation}\mathcal{G}i(t) = \left( \mu_{i}(t), R_{i}, s_{i}, \alpha_{i}, c_{i} \right)\end{equation}where $R_{i} \in \mathbb{SO}(3)$ is the rotation matrix, $s_{i} \in \mathbb{R}^3$ is the anisotropic scaling vector, $\alpha_{i} \in [0, 1]$ is the opacity, and $c_{i} \in \mathbb{R}^{k}$ denotes the spherical harmonic coefficients for view-dependent color. We further employ a static-dynamic composition strategy. For static background Gaussians, the center $\mu_{i}(t)$ remains constant. For dynamic object Gaussians, we assume local linear motion within a segment:\begin{equation}\mu_{i}(t) = \mu_{i, \text{init}} + v_{i} \cdot (t - T_s)\end{equation}where $v_i \in \mathbb{R}^3$ represents the regressed velocity vector. By using the frames $\{T_{s}, T_{s+1}\}$ as context, our ReconDrive framework regresses these spatio-temporal parameters to construct a unified 4D representation. The detailed architecture and training strategy follow in the subsequent sections.

\subsection{Backbone and Prediction Heads}
\label{sec:method-gs}
We employ a feature backbone to extract dense image tokens, which are processed by a dual-path architecture consisting of the Gaussian Center Prediction Head (GCPH) and the Gaussian Parameter Prediction Head (GPPH). Together, these heads regress spatial coordinates and appearance attributes to form the complete Gaussian parameters.

\paragraph{Feature Backbone.}
The input consists of multi-view images from two context frames ($2\times O$ images, where $O=6$ for the nuScenes dataset~\cite{caesar2020nuscenes}). The backbone follows the VGGT~\cite{wang2025vggt} architecture, utilizing two cascaded stages to extract and fuse spatio-temporal features:
\begin{itemize}[leftmargin=9pt]
    \item \textbf{Image Tokenization:} Each image is first downsampled to a resolution of $H' \times W'$ (with a scale factor of 3). We then utilize a DINOv2~\cite{oquab2023dinov2} encoder to transform the resized images into dense feature tokens. The encoder applies a patch size of $p=14$, resulting in tokens of dimension $d=1024$ and a spatial grid resolution of $(H' / p) \times (W' / p)$.
    \item \textbf{Alternating-Attention Fusion:} The tokens from all $2 \times O$ images are concatenated and passed through 24 Transformer layers utilizing Alternating-Attention. This mechanism alternates between frame-wise self-attention, which maintains local structural consistency within each timestamp, and global self-attention, which captures cross-view and temporal geometric correlations.
\end{itemize}

\paragraph{Hybrid Gaussian Prediction Head.} While VGGT~\cite{wang2025vggt} excels at recovering explicit 3D geometry, such as depth and point maps, its latent features are optimized for structural consistency rather than the fine-grained photometric details required for high-fidelity rendering. To bridge this gap, we move beyond direct regression from backbone features and design the Hybrid Gaussian Prediction Head to leverage the geometric priors of 3D foundation models while resolving their inherent limitations. Specifically, it overcomes photometric deficiency by fusing raw image cues into the attribute regression path for high-fidelity rendering. Simultaneously, it eliminates spatial misalignment by explicitly incorporating sensor calibration into the GCPH, ensuring the reconstructed scene is precisely anchored within the ego-vehicle’s calibrated coordinate system. In the Appendix, we further illustrate the spatial misalignment encountered when directly employing VGGT for autonomous driving scene reconstruction and 3D geometry generation.

\begin{itemize}[leftmargin=9pt]
    \item \textbf{Gaussian Center Prediction Head.} We firstly adopt a Dense Prediction Head (DPT)~\cite{ranftl2021vision} to upsample the fused transformer features back to the originally resized image resolution ($H^{'}\times W^{'}$) as $FC_{t}^{o}$. Each $FC_{t}^{o}$ is then input into a $3~\times 3$ convolutional layer to generate the pixel-wise depth map $DP_t^{o}$ with $H^{'}\times W^{'}$ size. Next, we project each depth map into the 3D space with camera intrinsic and extrinsic parameters, to obtain the gaussian 3D center corresponding to the pixel. Note that the two frames' gaussian 3D centers are located in their ego coordinate separately. Utilizing the calibrated sensor data improves more accurate gaussian location prediction.

    \item \textbf{Gaussian Parameter Prediction Head.} Similar to Gaussian Center Prediction Head, we also use DPT to upsample the fused transformer features back to the originally resized image resolution ($H^{'}\times W^{'}$) as $F_{t}^{o}$. Then we incorporate a shortcut connection to concatenate the original image with the upsampled features to $FP_{t}^{o}$. This is extremely important in the gaussian parameters prediction, because it enables the capture of high-frequency texture and color details that may be lost in transformer feature downsampling. A convolutional layer is used to process $FP_{t}^{o}$ to generate the extra gaussian parameters.
\end{itemize}

Finally, we obtain a pixel-wise 3D gaussian set of the two context frames $T_s$ and $T_{s+1}$: $\mathcal{G}(T_s)$ and $\mathcal{G}(T_{s+1})$.

Furthermore, to mitigate the domain gap between general 3D data and large-scale urban driving scenes, we adopt a parameter-efficient fine-tuning strategy. We utilize frozen pre-trained VGGT~\cite{wang2025vggt} weights as a robust structural prior and apply LoRA~\cite{hu2022lora} to adapt the model to the autonomous driving domain. For our implementation, we configure the LoRA adapters with a rank $r=8$ and a scaling factor $\alpha=32$.

\subsection{4D Gaussian Generation}
\label{sec:method-4dgs}
This part further explains how we obtain 4D Gaussians with static-dynamic composition and generated $ \mathcal{G}_{T_{s}}$ and $ \mathcal{G}_{T_{s+1}}$.

\paragraph{Dynamic Object Mask and Motion Estimation.} The autonomous driving environment can be primarily categorized into two components: static background elements (e.g., sky, roads, buildings, and vegetation) and dynamic traffic participants (e.g., vehicles and pedestrians)—the latter being the main contributors to dynamic scene changes. Here, we estimate the motion flow for the Gaussian kernels corresponding to these dynamic traffic participants.

First, we leverage the foundation model SAM2~\cite{ravi2024sam} to extract instance-level regions of traffic participants (with a particular focus on vehicles) from the two context frames. SAM2 is chosen for its robust mask extraction capability and efficient inference speed, which aligns with our real-time inference requirements. 
Next, we calculate the velocities of dynamic objects in the ego-coordinate system of frame $T_{s}$. The nuScenes dataset~\cite{caesar2020nuscenes} provides 3D bounding box annotations for each object. For each dynamic object, we transform its 3D location at frame $T_{s+1}$ into the ego-coordinate system of frame $T_{s}$, then compute the object’s displacement. Within the short time segment $[T_{s}, T_{s+1}]$, we assume the object undergoes rigid-body motion with a constant velocity. Thus, the object’s velocity is estimated as:
\begin{equation}
    v = \textit{displacement}~/~(T_{s+1} - T_{s})
\end{equation}
\begin{wrapfigure}{r}{0.42\textwidth}
\vspace{-8pt}
\centering
\includegraphics[width=0.40\textwidth]{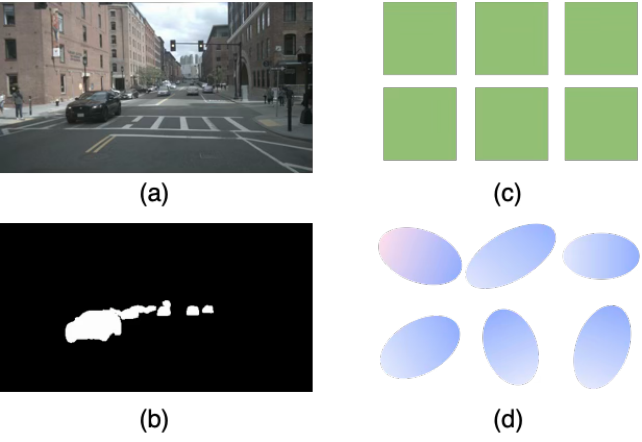}
\vspace{-10pt}
\caption{\textbf{Mapping Among Images, Masks, Dense Features, and Gaussian Indices.}
(a) Input image. (b) Dynamic object mask.
(c) Dense features. (d) Gaussian kernels.
Each pixel in the dense feature map generates a corresponding Gaussian kernel,
enabling consistent pixel-wise mapping across all these components.}
\vspace{-10pt}
\label{fig:mask-gaussians-mapping}
\end{wrapfigure}
Note that we can also calculate the object displacement by two-frame gaussian centers for dataset without 3d object annotations.
Through this process, we obtain the dynamic object motion flow $V$ with dimensions $H^{'} \times W^{'} \times 3$, where all pixel regions within the object’s mask share the same velocity. As illustrated in Fig.~\ref{fig:mask-gaussians-mapping}, a pixel-wise spatial mapping exists between the image space and the dense feature space associated with the Gaussian kernels. Thus, for the $i$-th Gaussian kernel at frame $T_{s}$, we derive its 3D center at any time $t$ using the following equation:
\begin{equation}
    \mu_{i}(t) = \mu_{i}(T_{s}) + V(i) \cdot (t - T_{s}), \quad t \in [T_{s}, T_{s+1}].
\end{equation}

\paragraph{Temporal Alignment and Fusion.}
Here we explain how we align and fuse the two context-frame Gaussians $\mathcal{G}(T_s)$ and $\mathcal{G}(T_{s+1})$ into final 4D Gaussians for segment $[T_{s}, T_{s+1}]$ by transforming the frame $T_{s}$'s Gaussians into frame $T_s$'s ego coordinate and time $T_s$. Firstly, we spatially transform the frame $T_{s+1}$'s Gaussians into frame $T_{s}$'s ego coordinate system with the ego transformation matrix from frame $T_i$'s ego vehicle pose $E_{T_{i}}$ to frame $T_{i+1}$'s ego vehicle pose $E_{T_{i+1}}$. We mainly consider transforming the gaussian centers, rotation, and color coefficients parameters.  Then, temporally align the $\mathcal{G}(T_{s+1})$ to time $T_s$ by moving the dynamic object gaussian centers with velocity flow $V$. The two steps cannot be swapped because the velocity field is computed in $T_{s}$'s ego frame. Finally, we get transformed Gaussians $\mathcal{G}^{'}(T_{s+1})$, and concatenate the two Gaussians $\mathcal{G}(T_{s})$ and $\mathcal{G}^{'}(T_{s+1})$. These concatenated Gaussians and the velocity flow comprise the final 4D gaussian splatting.

\subsection{Training}
\label{sec:method-training}
\paragraph{Training Loss.}
To enhance visual fidelity, we adopt perceptual loss using a pre-trained VGG-19 network~\cite{simonyan2014very}, which computes the L1 distance between high-level features of rendered images $\hat{I}$ and real images $I$:
$\mathcal{L}_{\text{percep}} = \left\| \phi(\hat{I}) - \phi(I) \right\|_1$,
where $\phi(\cdot)$ denotes the feature extractor of a specific convolutional layer, encouraging perceptually consistent features. L2 loss is used to constrain pixel-wise intensity differences between rendered and real images: $\mathcal{L}_{l2} = \left\| \hat{I} - I \right\|_2$. Combining them, the rendering loss for supervising photometric rendering is formulated as:
\begin{equation}
\mathcal{L}_{\text{render}} = \lambda_{\text{percep}} \cdot \mathcal{L}_{\text{percep}} + \lambda_{l2} \cdot \mathcal{L}_{l2}
\end{equation}

To stabilize Gaussian center prediction training and improve geometric accuracy, we introduce projection loss to enforce consistency when warping a single source frame $t$ to the reference frame $t'$. It combines a weighted L1 loss (measuring pixel-wise intensity differences between warped predicted images $\hat{I}_{\text{warped}}$ and warped ground-truth images $I_{\text{warped, GT}}$) and a weighted SSIM loss (evaluating structural and perceptual consistency via $1 - \text{SSIM}(\hat{I}_{\text{warped}}, I_{\text{warped, GT}})$). Both components are computed under a valid warping mask, with the final loss defined as:
\begin{equation}
\mathcal{L}_{\text{project}} = \mathcal{L}_{\text{masked}} \left( \lambda_{l1} \cdot \mathcal{L}_{l1} + \lambda_{\text{ssim}} \cdot \mathcal{L}_{\text{ssim}} \right)
\end{equation}
where $\mathcal{L}_{\text{masked}}$ is the mask-weighted loss computation, and $\lambda_{l1}, \lambda_{\text{ssim}}$ are the weight coefficients for L1 and SSIM loss.
More projection loss details are provided in the Appendix.

The norm loss acts as a regularization term for Gaussian scale and opacity parameters:
\begin{equation}
\mathcal{L}_{\text{norm}} = \lambda_{\text{scale}} \cdot \mathbb{E}\left[ \|\text{scale}_\text{maps}\|_2 \right] + \lambda_{\text{opacity}} \cdot \mathbb{E}\left[ |\text{opacity}_\text{maps}| \right]
\end{equation}
where $\text{scale}_\text{maps}$ denotes Gaussian scale parameter maps, and the L2-norm $\|\cdot\|_2$ constrains scale magnitude to avoid extreme values; $\text{opacity}_\text{maps}$ represents Gaussian opacity maps, and the L1-norm enforces sparsity by encouraging opacities to converge to 0 or 1. $\lambda_{\text{scale}}$ and $\lambda_{\text{opacity}}$ are corresponding weight coefficients.

Finally, ReconDrive is trained end-to-end with the total loss:
\begin{equation}
\mathcal{L} = \mathcal{L}_{\text{render}} +  \mathcal{L}_{\text{project}} + \mathcal{L}_{\text{norm}}.
\end{equation}~\label{eq:training loss}

\section{Experiments}
\label{sec:experi}

\begin{figure*}[t]
    \setlength{\belowcaptionskip}{0pt}
	\centering
	\includegraphics[width=1.0\textwidth]{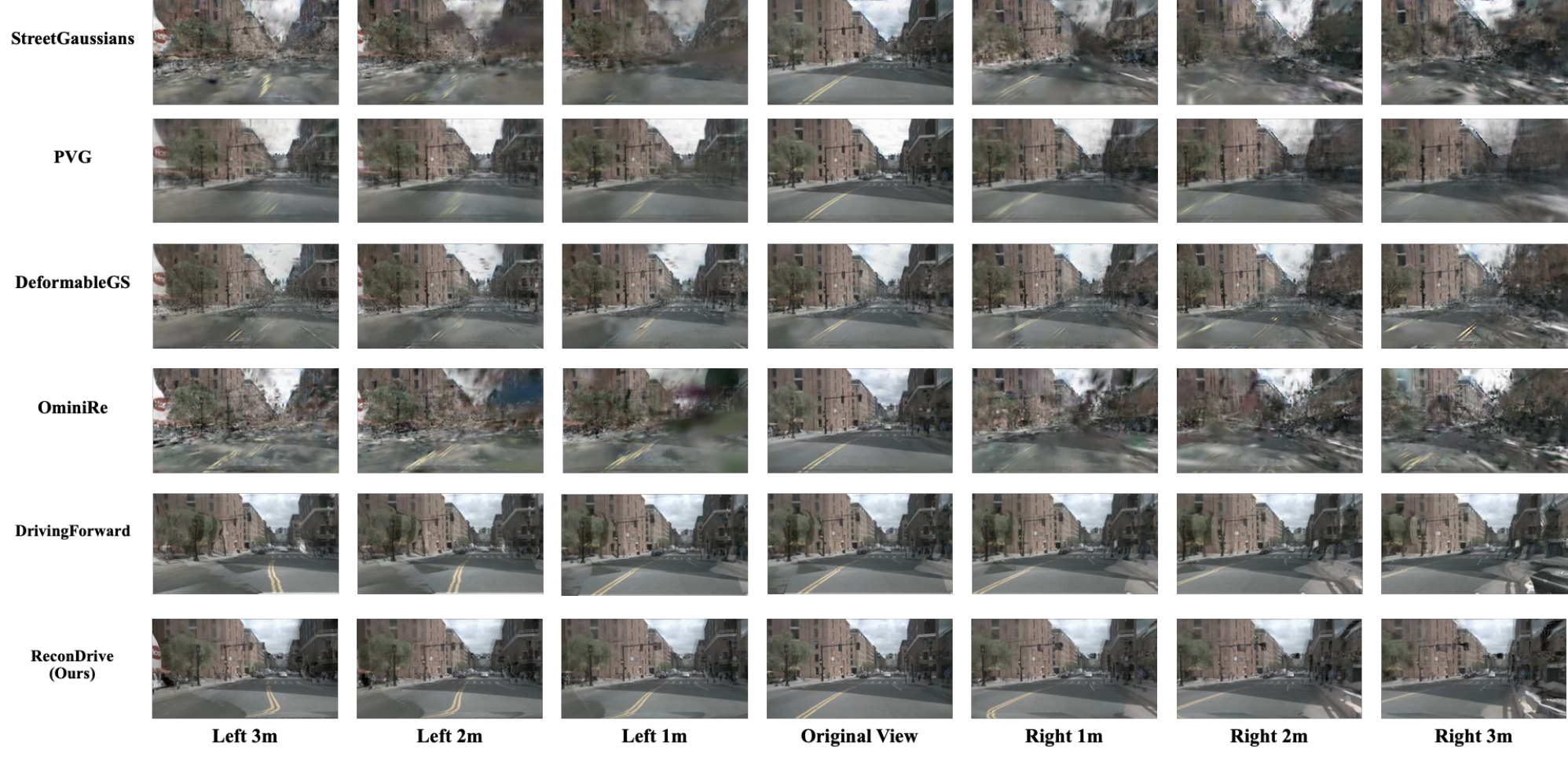}
	\caption{\textbf{Visual Comparisons of Scene Reconstruction and Novel-View Synthesis.}  
    Compared with per-scene optimization methods (Street Gaussians, PVG, DeformableGS, OminiRe), our ReconDrive maintains high-quality visual rendering in both scene reconstruction (Original View) and novel-view synthesis (1–3 meters of lateral movement leftward and rightward). Compared with the feed-forward method DrivingForward, ReconDrive preserves accurate geometric consistency—evident in the tree and the surroundings—and the performance improvement becomes more pronounced as the moving distance increases. Additionally, it exhibits less image distortion and blurriness, particularly in image boundary regions.
    Extended visual comparisons are provided in the Appendix.
    }\label{fig:comparison-vis-1}
\end{figure*}

This section presents a visual urban scene reconstruction benchmark comprising state-of-the-art per-scene optimization and feed-forward methods on the nuScenes~\cite{caesar2020nuscenes} dataset, and compares our approach against these baselines under multiple evaluation protocols to demonstrate its effectiveness.
\begin{table*}[htbp]
\centering
\renewcommand\arraystretch{1.0}
\resizebox{\linewidth}{!}{
\begin{tabular}{lccccccc}
\toprule
		\multirow{2}{*}{Method} &
		\multicolumn{3}{c}{Visual Scene Reconstruction} & 
		\multicolumn{3}{c}{Novel-View Synthesis} &
        Inference Speed~$\downarrow$ \\
		& PSNR~$\uparrow$ & SSIM~$\uparrow$ & LPIPS~$\downarrow$ & PSNR~$\uparrow$ & SSIM~$\uparrow$ & LPIPS~$\downarrow$ & (Per Scene) \\
\midrule
\multicolumn{8}{l}{\textit{Per-Scene Optimization Methods}} \\
Street Gaussians~\cite{yan2024street} & 29.18 & 0.8824 & 0.1658 & 22.98 & 0.6959 & 0.2948 &  31min \\
PVG~\cite{chen2023periodic} & 29.58 & 0.8839 & 0.2200 & 23.48 & 0.6919 & 0.2897 & 23min \\
DeformableGS~\cite{yang2024deformable} & 28.93 & 0.8832 & 0.1610 & 23.73 & 0.6919 & \textbf{0.2342} & 46min \\
OminiRe~\cite{chen2025omnire} & 29.42 & 0.8853 & 0.1577 & 23.01 & 0.6885 & 0.2762 & 35min \\

\hline
\multicolumn{8}{l}{\textit{Feed-Forward Methods}} \\
Drivingforward~\cite{tian2025drivingforward} & 22.83 & 0.7650 & 0.2563 & 21.88 & 0.6866 & 0.2979 & 5s \\
\textbf{ReconDrive (Ours)} & \textbf{32.66} & \textbf{0.9589} & \textbf{0.0618} & \textbf{23.99} & \textbf{0.7234} & 0.2591 & 15s \\
\bottomrule
\end{tabular}
}
\caption{\textbf{Scene Reconstruction and Novel-View Synthesis Evaluation Results on the nuScenes Dataset.}}
\label{tab:recons and nvs evaluation.}
\end{table*}

\subsection{Benchmark}
\paragraph{Dataset.}
The nuScenes~\cite{caesar2020nuscenes} is a classical autonomous driving dataset originally designed for 3D perception tasks, comprising 1,000 urban scenes. Each scene has approximately 20 s duration and contains multi-view images captured by six cameras. We use the original 700 training scenes to train the feed-forward model and select 14 representative validation scenes from the original validation split, covering diverse conditions in time (day/night), weather (sunny/rainy), driving behaviors (static/straight/turning), and traffic density. The detailed list of selected validation scenes is provided in the Appendix. Following~\cite{li2025uniscene}, we adopt a frame rate of 12 Hz as key frames. For each evaluation scene, frames at 0.5-second intervals (i.e., every 6th frame) serve as context frames for reconstruction, while the remaining frames are reserved for novel-view synthesis evaluation.

\paragraph{Evaluation Protocols and Metrics.}
We evaluate the performance of models in the following three manners. We also report the gaussian generation speed for each scene.
\begin{itemize}[leftmargin=9pt]
    \item Visual Scene Reconstruction. We use the context frames (i.e., every 6th frame) as ground truth images, and use Peak Signal-to-Noise Ratio (PSNR), Structural Similarity Index Mapping (SSIM), and Learned Perceptual Image Patch Similarity (LPIPS) as evaluation metrics.
    \item Novel-View Synthesis. We use the non-context frames (i.e., every 2nd-5th frames) as ground truth images, and also use PSNR, SSIM, and LPIPS as evaluation metrics.
    \item Novel-View Synthesis for 3D Perception. To evaluate novel-view synthesis for 3D perception, we simulate lateral camera shifts by displacing the ego-vehicle trajectory by $\pm 1$m, $\pm 2$m, and $\pm 3$m relative to the original route. We then evaluate the 3D perception performance of UniAD~\cite{hu2023planning}, which is pre-trained on the original nuScenes images at 2 Hz, using these rendered multi-view outputs across all seven offsets ($0$m and $\pm 1$m, $\pm 2$m, $\pm 3$m). For a consistent evaluation, we also provide the rendered images to UniAD at a 2 Hz frame rate. 3D perception results include 3D object detection and tracking (only vehicle type).
\end{itemize}

\paragraph{Baseline Models and Settings.}
We evaluate two categories of baseline approaches, comprising per-scene optimization methods and feed-forward methods. For the optimization-based category, we adapt the open-source DriveStudio~\cite{chen2025omnire} codebase to support the 12 Hz nuScenes dataset~\cite{caesar2020nuscenes}, as it was originally designed for the Waymo dataset~\cite{sun2020scalability}. Within this framework, we evaluate four state-of-the-art optimization methods: Street Gaussians~\cite{yan2024street}, PVG~\cite{chen2023periodic}, DeformableGS~\cite{yang2024deformable}, and OmniRe~\cite{chen2025omnire}. Each model is trained on the designated context frames for 30,000 iterations per validation scene. In the feed-forward category, we compare our work against the state-of-the-art DrivingForward~\cite{tian2025drivingforward} approach. To ensure a fair comparison under similar settings, we provide multiple frames as input to this baseline. All methods are evaluated at a resolution of $518 \times 280$ pixels, with detailed results provided in Table~\ref{tab:recons and nvs evaluation.} and Table~\ref{tab:perception-evaluation.}.

\subsection{Main Experiment Results}
\paragraph{Implementation Details.} We set the training-loss hyperparameters to $\lambda_{\text{percep}}=0.05$, $\lambda_{l2}=1.0$, $\lambda_{l1}=0.85$, $\lambda_{\text{ssim}}=0.15$, and $\lambda_{\text{scale}}=\lambda_{\text{opacity}}=0.01$. The model is optimized with AdamW using gradient accumulation (8 steps). Training proceeds in two stages: (1) Single-frame pre-training for 10 epochs (batch size 4, learning rate $2\times10^{-5}$, weight decay 0.01), where each input frame $T_i$ is also used as ground truth; (2) Dual-frame fine-tuning for 2 epochs (batch size 2). Dynamic-object masks and motion are extracted for five vehicle classes (car, truck, bus, trailer, construction vehicle). All experiments run on H800 GPUs, requiring ~8 GPU-days. Evaluation results are reported in Tab.~\ref{tab:recons and nvs evaluation.} and Tab.~\ref{tab:perception-evaluation.}.
Figure \ref{fig:comparison-vis-1} presents visual comparisons of scene reconstruction and novel-view synthesis using Scene-0103 from the nuScenes validation set. These qualitative results demonstrate ReconDrive's ability to achieve high-fidelity rendering while maintaining precise spatial alignment during view synthesis.

\paragraph{Visual Scene Reconstruction.} As shown in Tab.~\ref{tab:recons and nvs evaluation.}, per-scene optimization methods exhibit strong performance due to their iterative parameter tailoring. Specifically, PVG~\cite{chen2023periodic} and OminiRe~\cite{chen2025omnire} achieve the highest PSNR (29.58) and the highest SSIM (0.8853) respectively, while DeformableGS~\cite{yang2024deformable} lead in LPIPS. In contrast, existing feed-forward baselines~\cite{tian2025drivingforward} lag significantly (e.g., PSNR 22.83 for Drivingforward), highlighting the traditional performance gap between generalizable and scene-specific reconstruction.
ReconDrive (Ours) effectively bridges this gap. It outperforms \textit{all} baselines in pixel, structural and perceptual metrics, achieving a superior PSNR of 32.66 and SSIM of 0.9589 and LPIPS of 0.0618. 

\paragraph{Novel-View Synthesis Evaluation.} As shown in Tab.~\ref{tab:recons and nvs evaluation.}, per-scene optimization methods generally outperform prior feed-forward baselines in novel-view synthesis, as iterative parameter tuning better captures complex view-dependent variations. For instance, DeformableGS~\cite{yang2024deformable} and Street Gaussians~\cite{yan2024street} achieve PSNRs of 23.73 and 22.98, respectively, while the feed-forward Drivingforward~\cite{tian2025drivingforward} reaches only 21.88.
ReconDrive effectively bridges the performance gap between efficiency and quality, establishing a new state-of-the-art for feed-forward 4DGS. While maintaining a comparable perceptual quality (LPIPS) to the top-performing baseline, ReconDrive notably surpasses DeformableGS~\cite{yang2024deformable} by 0.26 dB in PSNR and 0.0315 in SSIM.. Furthermore, ReconDrive demonstrates substantial gains over existing feed-forward approaches, exceeding Drivingforward by 2.11 dB in PSNR and 0.0368 in SSIM, while significantly reducing perceptual error (LPIPS) by 0.0388. These results underscore that ReconDrive not only eliminates the traditional trade-off between inference speed and synthesis quality but also provides superior structural and perceptual consistency for novel-view generation in complex urban scenes.

\paragraph{3D Perception Evaluation.}
\begin{wraptable}{r}{0.52\textwidth}
\vspace{-8pt}
\centering
\renewcommand\arraystretch{1.0}

\vspace{-6pt}
\resizebox{1.0\linewidth}{!}{
\begin{tabular}{lcc}
\toprule
\multirow{2}{*}{Method} &
\multicolumn{1}{c}{Detection} & 		\multicolumn{1}{c}{Tracking} \\
& mAP (\%)~$\uparrow$  & AMOTA (\%)~$\uparrow$ \\
\midrule
\multicolumn{3}{l}{\textit{Per-Scene Optimization Methods}} \\
Street Gaussians~\cite{yan2024street} & 14.6 & 11.9 \\
PVG~\cite{chen2023periodic} & 18.5 & 14.4 \\
DeformableGS~\cite{yang2024deformable} & 16.4 & 13.4 \\
OminiRe~\cite{chen2025omnire} & 16.1 & 12.9 \\
\hline
\multicolumn{3}{l}{\textit{Feed-Forward Methods}} \\
Drivingforward~\cite{tian2025drivingforward} & 23.4 & 13.3 \\
\textbf{ReconDrive (Ours)} & \textbf{26.7} & \textbf{18.9} \\
\bottomrule
\end{tabular}
}
\caption{\textbf{3D Perception Results on Novel-View Rendered Scenes with 0 m, ±1 m, ±2 m, and ±3 m View Offsets.}}
\vspace{-6pt}
\label{tab:perception-evaluation.}
\end{wraptable}

Analysis of the 3D perception results in Table 2 reveals that per scene optimization methods, such as Street Gaussians~\cite{yan2024street}, exhibit limited efficacy with mAP $\le$ 14.6\% and AMOTA $\le$ 11.9\%. While feed forward approaches generally excel in detection—evidenced by DrivingForward~\cite{tian2025drivingforward} achieving an mAP of 23.4\%—they often fall short in tracking consistency, as seen in the AMOTA of 13.3\% for DrivingForward compared to 14.4\% for PVG~\cite{chen2023periodic}. In contrast, ReconDrive achieves state of the art performance across all metrics, reaching 26.7\% mAP and 18.9\% AMOTA. These results underscore ReconDrive’s robust 3D perception capabilities, demonstrating superior detection accuracy and tracking consistency even when subjected to novel view offsets.

\paragraph{Inference Efficiency.} To evaluate scalability, we calculate the average time cost for Gaussian generation per scene ($\approx$20 s duration). In ReconDrive, as adjacent temporal segments share context frames, we use caching mechanism to eliminate redundant computation during inference. With this optimization, ReconDrive achieves an inference speed of 15s per scene—orders of magnitude faster than per-scene optimization methods~\cite{yan2024street, chen2023periodic}, which require about 30 minutes. While slightly slower than existing feed-forward baselines~\cite{tian2025drivingforward} (5s), ReconDrive offers a superior balance between high-fidelity reconstruction and computational efficiency.
\section{Conclusion}
\label{sec:conclu}
Developing efficient and scalable scene reconstruction methodologies is a fundamental requirement for building closed-loop evaluation and simulation platforms within the autonomous driving industry. In this paper, we introduced ReconDrive, a feed-forward framework for efficient 4D Gaussian Splatting generation specifically tailored for large-scale driving scenes. Unlike existing per-scene optimization methods that require extensive computational resources, our model generates a complete scene representation in a single forward pass. This efficiency is achieved through a static-dynamic composition strategy and segment-wise Gaussian aggregation, which effectively adapt a 3D foundation model via our specialized Gaussian prediction architecture.

Experimental results on our new nuScenes-based benchmark demonstrate that ReconDrive outperforms state-of-the-art per-scene optimization baselines across nearly all evaluation metrics. To the best of our knowledge, this work represents the first instance where a feed-forward approach surpasses optimization-based methods in reconstruction and novel-view synthesis performance through comprehensive experimentation. These findings provide a clear direction and renewed confidence for the development of scalable, generative simulation environments for large-scale autonomous driving.

\paragraph{Limitations and Future Work.} This study represents an initial effort toward achieving large-scale autonomous driving scene reconstruction and novel-view synthesis via feed-forward 3D foundation models. While our results demonstrate significant potential for this paradigm, several challenges remain to be addressed in future research.
\begin{itemize}
    \item Temporal Representation of Non-Rigid Motion: ReconDrive currently utilizes an explicit 4D Gaussian Splatting representation based on linear motion estimation. While efficient, this formulation may struggle to accurately represent complex non-rigid deformations or highly non-linear object trajectories. Future work could involve integrating more expressive temporal kernels or learnable deformation fields to better capture these dynamics.
    \item Temporal Consistency and Redundancy in Aggregation: Our current post-processing strategy for multi-frame aggregation relies on pixel-wise outputs that may lead to Gaussian redundancy and suboptimal handling of occluded regions. Transitioning toward a more integrated temporal fusion mechanism within the latent space could reduce redundancy and improve the structural integrity of occluded surfaces.
    \item Precision in Dynamic Object Extraction: The current pipeline relies on SAM2 for object segmentation, which can occasionally lead to inaccurate boundaries or missed detections. Furthermore, displacing dynamic objects directly can leave disocclusion artifacts in the background. Investigating joint inpainting techniques or end-to-end motion modeling could help mitigate these "background holes" and improve boundary fidelity.
    \item Computational Efficiency and Throughput: Although our feed-forward approach is orders of magnitude faster than optimization-based methods, there remains substantial room for architectural optimization. Researching lightweight foundation backbones and more efficient Gaussian sampling strategies will be critical for achieving real-time performance in edge-computing environments.
    \item Scalability and Generalization: Expanding the training scope to encompass larger, more diverse datasets is essential. Evaluating the model across a broader range of geographic locations and extreme weather conditions will be a primary focus to ensure the robustness and generalizability of the feed-forward paradigm. Broadening data diversity to cover different driving behaviors may further enhance robustness across varied real-world scenarios~\cite{hao2025styledrive}.
\end{itemize}

\clearpage
\bibliography{main}
\bibliographystyle{bibstyle}

\newpage
\setcounter{section}{0}
\appendix
\section*{Appendix}
    This appendix is organized as follows: Section~\ref{sec:appendix-validation} provides the full list of validation scenes. Section~\ref{sec:appendix-imple-details} contains additional implementation details. Section~\ref{sec: appendix-ablation-vggt} presents an ablation study on the spatial misalignment encountered when directly using VGGT~\cite{wang2025vggt} for 3D geometry generation. Section~\ref{sec:appendix-temporal-inputs} presents an ablation study evaluating how multi-frame inputs enhance the capability for novel-view synthesis. Finally, Section~\ref{sec:appendix-gaussian-flow} provides visualizations of dynamic Gaussians, and Section~\ref{sec:appendix-comparison} offers extensive qualitative comparisons of urban scene reconstruction and novel-view synthesis against baseline methods.

\section{Validation Scenes List}~\label{sec:appendix-validation}
In our experiment, we selected 14 representative scenes from the original nuScenes dataset’s~\cite{caesar2020nuscenes} validation split as the validation subset. Covering diverse conditions in time (day/night), weather (sunny/rainy), driving behaviors (static/straight/turning), and traffic density, these scenes effectively evaluate the novel-view synthesis ability for reconstruction while reducing the evaluation cost vs. the full validation split. The 14 selected scenes are: \textit{scene-0014, scene-0018, scene-0098, scene-0100, scene-0103, scene-0270, scene-0271, scene-0278, scene-0553, scene-0558, scene-0802, scene-0906, scene-0968, scene-1065}.

\section{Implementation Details}~\label{sec:appendix-imple-details}
\paragraph{Architecture.} 
We resize the original input images from the nuScenes dataset~\cite{caesar2020nuscenes} from a resolution of \(1600\times 900\times 3\) to \(518\times 280\times 3\). Following VGGT~\cite{wang2025vggt}, we adopt a \(14\times 14\) patch size and retain the camera token along with four register tokens in the feature backbone. Specifically, we feed the tokens from the 4-th, 11-th, 17-th, and 23-rd blocks into the Gaussian Center Prediction Head (GCPH) and Gaussian Parameter Prediction Head (GPPH), while excluding the camera token and four register tokens from the head inputs. For GCPH, we clamp outlier depth values to restrict the depth maps to a range of 1.5m (minimum) to 110m (maximum). For GPPH, we upsample the features to the resized image resolution (consistent with GCPH), process the resized image using a convolutional and ReLU layer, and then concatenate the resulting features with the upsampled features.

\begin{figure}[t]
    \setlength{\belowcaptionskip}{0pt}
	\centering
	\includegraphics[width=0.45\textwidth]{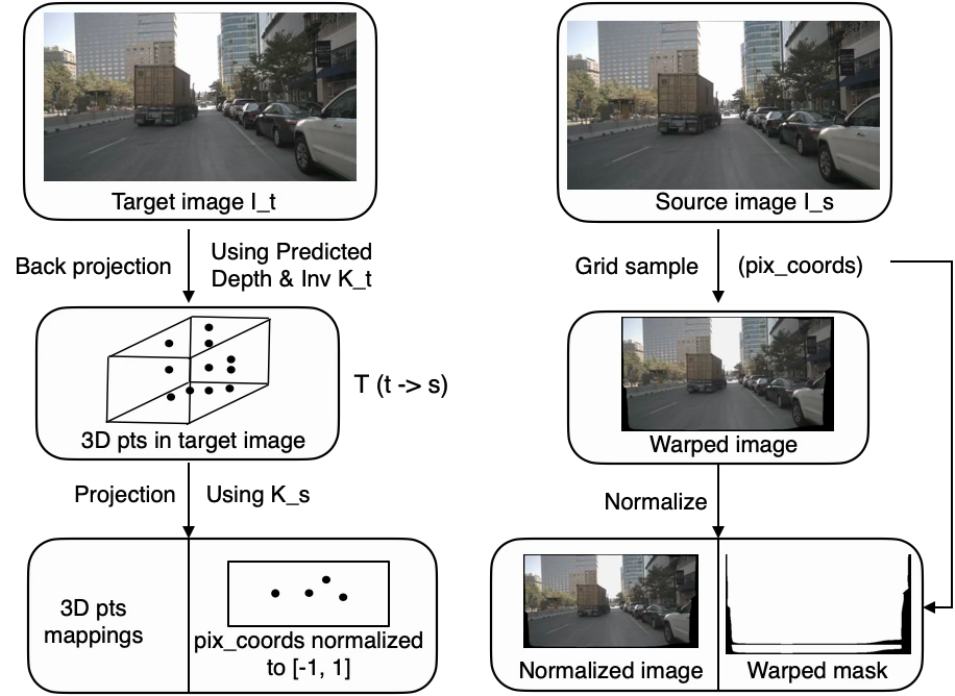}
	\caption{\textbf{Project Loss Computing Pipeline.} Note that \(K_t\) and \(K_s\) are identical since the camera is fixed.}\label{fig:project-loss}
\end{figure}

\paragraph{Training Loss (Project Loss).}
The project loss enforces photometric consistency between a warped source image and the target image, leveraging a two-stage pipeline to ensure geometric validity and appearance alignment for novel-view synthesis in urban scenes. Process is also illustrated in Fig.~\ref{fig:project-loss}.

Start with the Target Image \( I_t \) (captured at time \( t \)) and its corresponding predicted depth map \( \text{DP}_t \). First, backproject 2D pixels in \( I_t \) to 3D points in the target camera frame using the inverse camera intrinsics \( K_t^{-1} \) and \( \text{DP}_t \). For a pixel \( (u, v) \) in \( I_t \), the 3D point \( P_t \) is computed as:
\[
P_t = K_t^{-1} \cdot \begin{bmatrix} u \cdot D_t(u, v) \\ v \cdot D_t(u, v) \\ D_t(u, v) \end{bmatrix}
\]
Next, apply ego-motion transformation to convert \( P_t \) from the target frame to the source frame using the 6-DoF pose \( T(t \to s) \) (rigid transformation matrix), yielding \( P_s = T(t \to s) \cdot P_t \). Finally, perform projection + normalization to project \( P_s \) back to 2D pixel coordinates in the source camera view using the source camera intrinsics \( K_s \), then normalize the coordinates to the range \([-1, 1]\) (required for grid sampling) to get \(\text{pix\_coords}\).

Use \(\text{pix\_coords}\) from the geometric pipeline to warp the Source Image \( I_s \) via the grid\_sample operation (bilinear interpolation) in deep learning frameworks, generating the Warped Image \( I_s' \). Concurrently, generate a Warped Mask \( M_s\) to filter invalid regions:
\[
M_s = \begin{cases} 
1 & \text{if } \text{pix\_coords} \in [-1, 1]^2 \text{ and non-NaN} \\
0 & \text{otherwise}
\end{cases}
\]
This mask excludes pixels projected outside the source image bounds or occluded during transformation.

The Project Loss is defined as the masked L1 loss between the Warped Image \( I_s' \) and the Target Image \( I_t \), ensuring only valid pixels contribute to training:
\[
\frac{1}{\sum M_s + \epsilon} \sum_{u, v} \left| I_s'(u, v) \cdot M_s(u, v) - I_t(u, v) \cdot M_s(u, v) \right|
\]
where \( \epsilon = 10^{-8} \) is a small constant to avoid division by zero, and \( \sum M_s\) is the number of valid pixels in the Warped Mask. Element-wise multiplication with \( M_s\) suppresses gradients from invalid/occluded regions, enforcing geometric and photometric alignment. The final project loss combines a weighted L1 loss (measuring pixel-wise intensity differences between warped predicted images and warped ground-truth images) and a weighted SSIM loss (evaluating structural and perceptual consistency) as follows.
\begin{equation}
\mathcal{L}_{\text{project}} = \mathcal{L}_{\text{masked}} \left( \lambda_{l1} \cdot \mathcal{L}_{l1} + \lambda_{\text{ssim}} \cdot \mathcal{L}_{\text{ssim}} \right)
\end{equation}

\paragraph{Training.} To form training batches, we segment each scene into consecutive 6-frame clips, each covering a 0.5-second duration, with the 1st and 6th frames serving as context frames. During training, for Project Loss computation, we designate the 1st frame as the target frame \( I_t \) and the subsequent frame (i.e., the 2nd frame) as the source frame \( I_s \). Additionally, we render the 4D Gaussians into different time frames, including novel-time frames (spanning from the 1st to the 5th frame), with corresponding sampling probabilities of [0.7, 0.3, 0.2, 0.1, 0.1, 0.05].

\section{Ablation Study on Spatial Misalignment with Original VGGT}\label{sec: appendix-ablation-vggt}
\begin{figure}[t]
    \setlength{\belowcaptionskip}{0pt}
	\centering
	\includegraphics[width=0.5\textwidth]{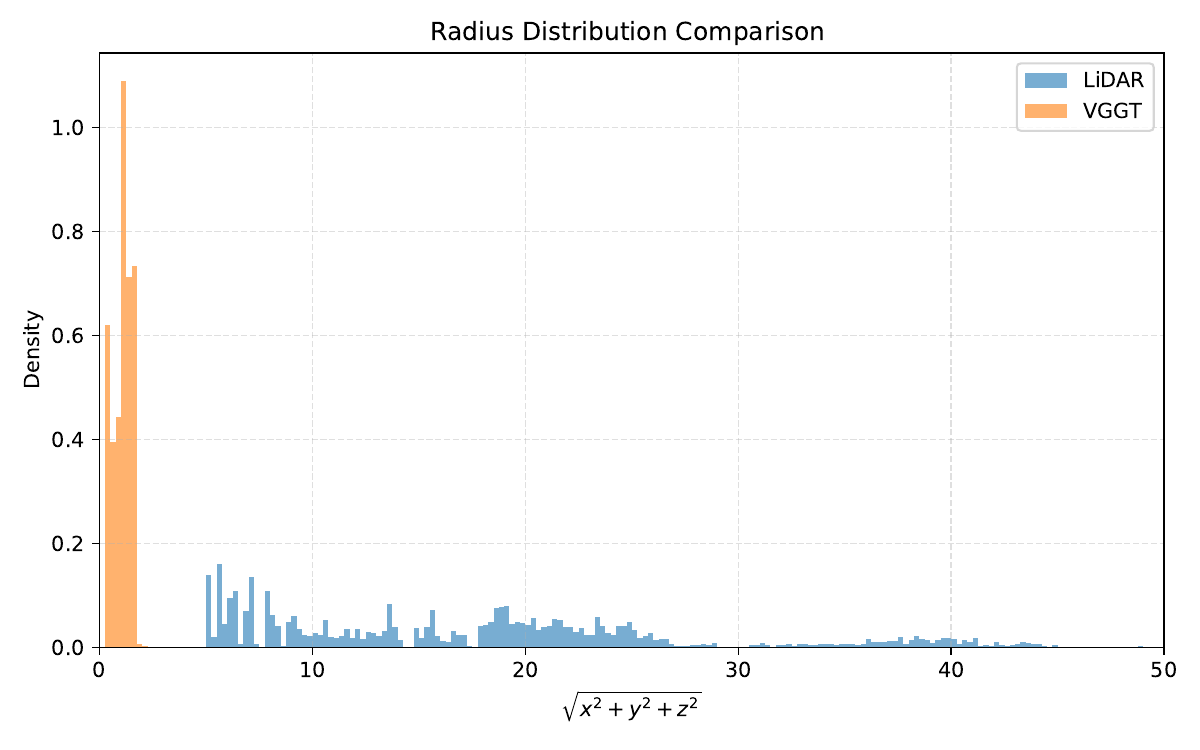}
	\caption{\textbf{Distribution of Spatial Distances between the Point Maps Generated by the Original VGGT and Ground Truth LiDAR Point Clouds}, highlighting significant metric misalignment.}\label{fig:appendix_vggt_distance_hist}
\end{figure}

In this section, we analyze the 3D geometric accuracy when directly employing the original VGGT model to process multi-view imagery. Specifically, we input a single frame consisting of six-view images into the original VGGT to estimate camera poses, depth maps, and point maps. We then evaluate the spatial alignment by calculating the distance distribution between the point maps generated by VGGT and the ground truth LiDAR point clouds from the nuScenes dataset. The resulting distribution is illustrated in Fig.~\ref{fig:appendix_vggt_distance_hist}.

\begin{figure*}[t]
    \setlength{\belowcaptionskip}{0pt}
	\centering
	\includegraphics[width=1.0\textwidth]{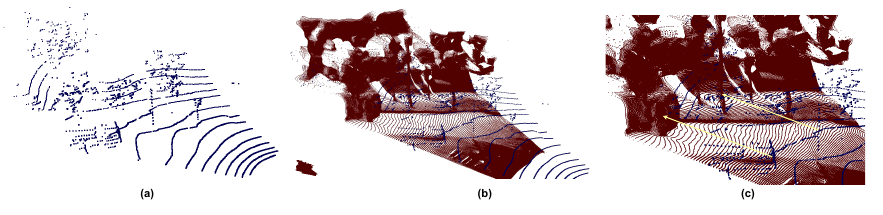}
	\caption{\textbf{Qualitative Comparison of VGGT-Generated Point Maps Across Various Scale Factors}, demonstrating the structural consistency despite metric scale ambiguity.}\label{fig:appendix_vggt_scales}
\end{figure*}

Our results reveal a significant scale misalignment between the original VGGT point map outputs and the LiDAR ground truths. This suggests that without domain-specific adaptation, the original VGGT cannot accurately reconstruct the scale of complex urban environments. However, as shown in the visualization in Fig.~\ref{fig:appendix_vggt_scales}, we observe that when a global scale factor is applied to the VGGT point maps, they align closely with the LiDAR data. This indicates that while the foundation model successfully captures the structural geometry of the scene, it fails to recover the absolute metric scale. These findings further underscore the necessity of incorporating pre-calibrated sensor parameters into the Gaussian prediction heads to build metrically accurate 4D Gaussian representations.

\section{Ablation Study for Temporal Inputs}\label{sec:appendix-temporal-inputs}
\begin{wraptable}{r}{0.52\textwidth}
\vspace{-8pt}
\centering
\renewcommand\arraystretch{1.0}

\vspace{6pt}

\resizebox{1.0\linewidth}{!}{
\begin{tabular}{lccc}
\toprule
\multirow{2}{*}{Method} &
\multicolumn{3}{c}{Novel-View Synthesis} \\
& PSNR~$\uparrow$ & SSIM~$\uparrow$ & LPIPS~$\downarrow$ \\
\midrule
ReconDrive-S & 23.54 & 0.6940 & 0.3177 \\
ReconDrive   & 23.99 (\textcolor{red}{+0.45}) & 0.7234 (\textcolor{red}{+0.0294}) & 0.2591 (\textcolor{red}{-0.058}) \\
\bottomrule
\end{tabular}
}
\captionof{table}{\textbf{Evaluation Results with Different Frame Input.}}

\vspace{-10pt}
\label{tab:ablation-temporal}
\end{wraptable}

To analyze how multi-frame inputs improve novel-view synthesis performance, we compare ReconDrive-S (single-frame input) against ReconDrive (two-frame input). As shown in Table~\ref{tab:ablation-temporal}, ReconDrive achieves a PSNR of 23.99, an SSIM of 0.7234, and an LPIPS of 0.2591, significantly outperforming ReconDrive-S across all metrics. This demonstrates that multi-frame fusion enhances synthesis quality because leveraging multiple frames expands the reconstructed visual field and provides complementary viewpoint information. Consequently, this leads to substantially higher fidelity and perceptual quality in the rendered outputs.

\section{Dynamic Gaussian Rendering Visualization}\label{sec:appendix-gaussian-flow}
\begin{figure*}[t]
    \setlength{\belowcaptionskip}{0pt}
	\centering
	\includegraphics[width=1.0\textwidth]{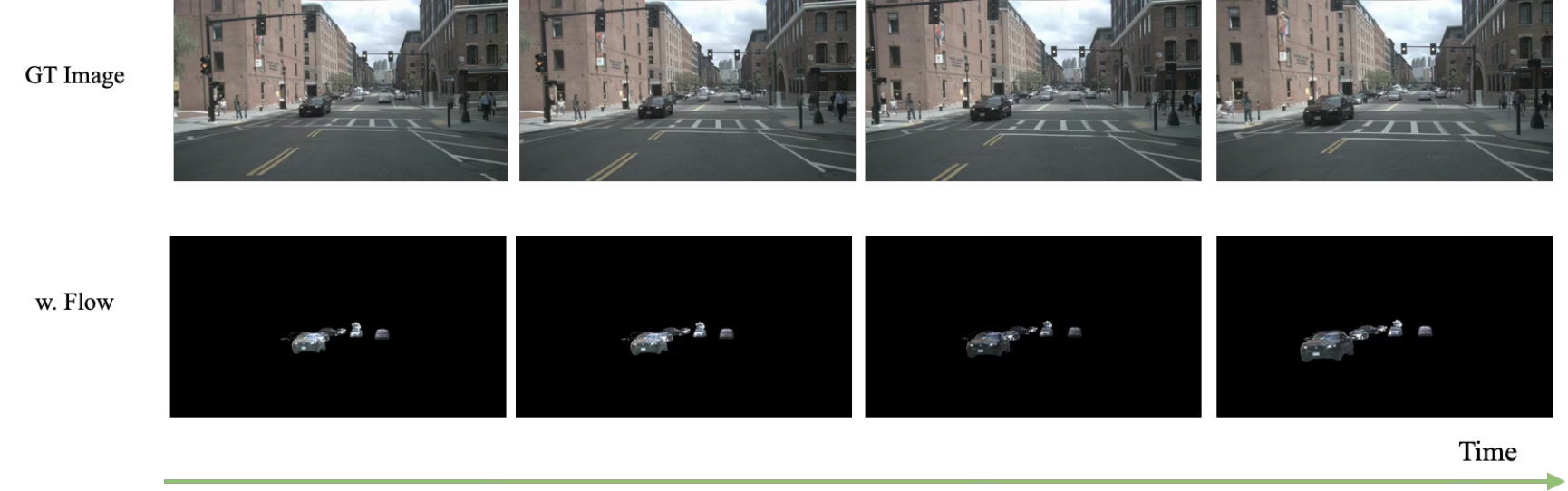}
	\caption{\textbf{Dynamic Gaussian Rendering Visualization.} The rendered dynamic vehicles exhibit accurate motion and align closely with their counterparts in the ground truth images. Note that we have enhanced the exposure of the rendered images for visualization purposes to facilitate comparative observation.}\label{fig:dynamic-gaussian}
\end{figure*}

In Fig.~\ref{fig:dynamic-gaussian}, we render the dynamic Gaussians as time progresses, masking out other static Gaussians. The resulting positions of the moving vehicles align well with the ground truth images, illustrating the effectiveness of our dynamic object mask and motion estimation for dynamic scene synthesis.

\section{Visualization Comparisons with Diverse Methods}\label{sec:appendix-comparison}
\begin{figure*}[t]
    \setlength{\belowcaptionskip}{0pt}
	\centering
	\includegraphics[width=1.0\textwidth]{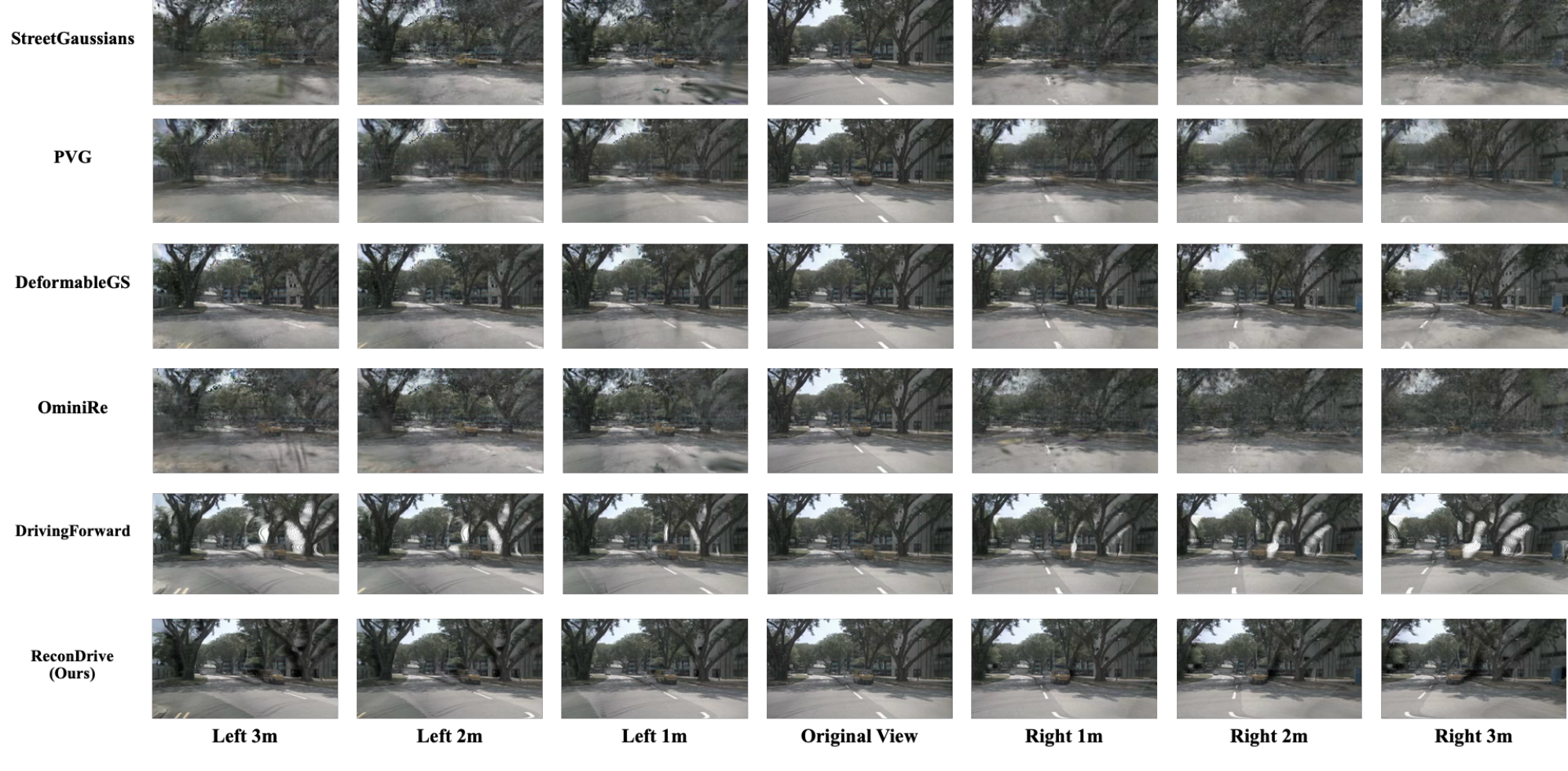}
	\caption{\textbf{Visual Comparisons of Scene Reconstruction and Novel-View Synthesis on Scene-0968 (from 14 Selected Validation Scenes).}
    Consistent with prior results, our ReconDrive maintains high-quality rendering in both scene reconstruction (Original View) and novel-view synthesis (1–3m lateral movement) compared to per-scene optimization methods (Street Gaussians, PVG, DeformableGS, OminiRe). Against the feed-forward DrivingForward, it preserves precise visual consistency—clear in tree near the car —with more notable performance gains as movement increases, plus less distortion/blurriness, especially at image boundaries.}\label{fig:comparison-vis-2}
\end{figure*}

\begin{figure*}[t]
    \setlength{\belowcaptionskip}{0pt}
	\centering
	\includegraphics[width=1.0\textwidth]{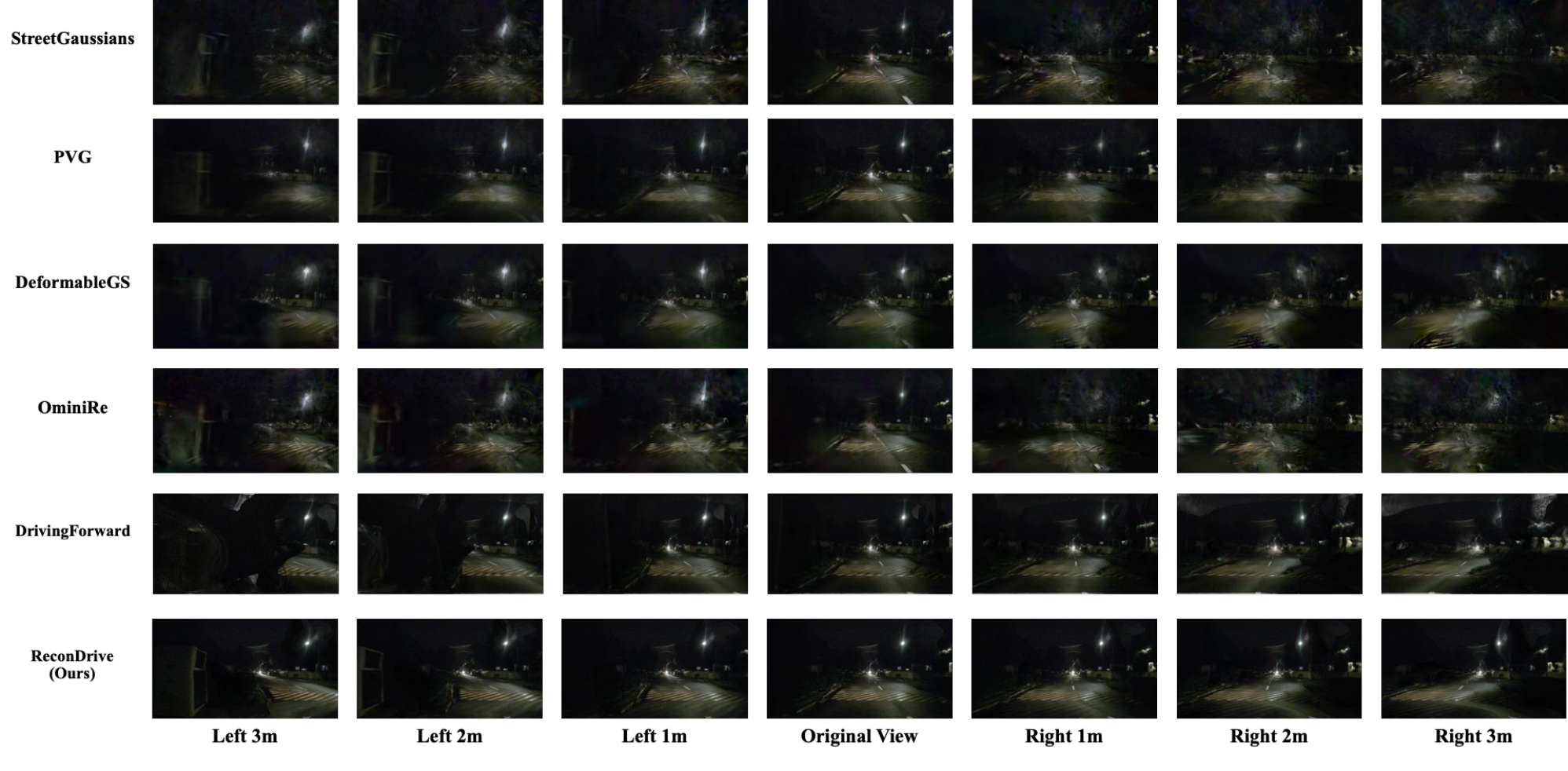}
	\caption{\textbf{Visual Comparisons of Scene Reconstruction and Novel-View Synthesis on Scene-1065 (from 14 Selected Validation Scenes).}
    Consistent with prior results, our ReconDrive maintains high-quality rendering in both scene reconstruction (Original View) and novel-view synthesis (1–3m lateral movement) compared to per-scene optimization methods (Street Gaussians, PVG, DeformableGS, OminiRe). Against the feed-forward DrivingForward (MF/SF modes), it preserves precise visual consistency. This visualization further demonstrates that ReconDrive performs exceptionally well in critical scenarios like nighttime environments.}\label{fig:comparison-vis-4}
\end{figure*}

To further demonstrate the effectiveness of our ReconDrive framework, we conduct extra visualization comparisons with five state-of-the-art methods, including Street Gaussians~\cite{yan2024street}, PVG~\cite{chen2023periodic}, DeformableGS~\cite{yang2024deformable}, OminiRe~\cite{chen2025omnire}, and DrivingForward~\cite{tian2025drivingforward}—under both Multi-Frame (MF) and Single-Frame (SF) modes across the selected validation scenes. The results (Figs.~\ref{fig:comparison-vis-2}–\ref{fig:comparison-vis-4}) demonstrate ReconDrive’s superiority in both scene reconstruction and novel-view synthesis.

\begin{figure*}[t]
    \setlength{\belowcaptionskip}{0pt}
	\centering
	\includegraphics[width=1.0\textwidth]{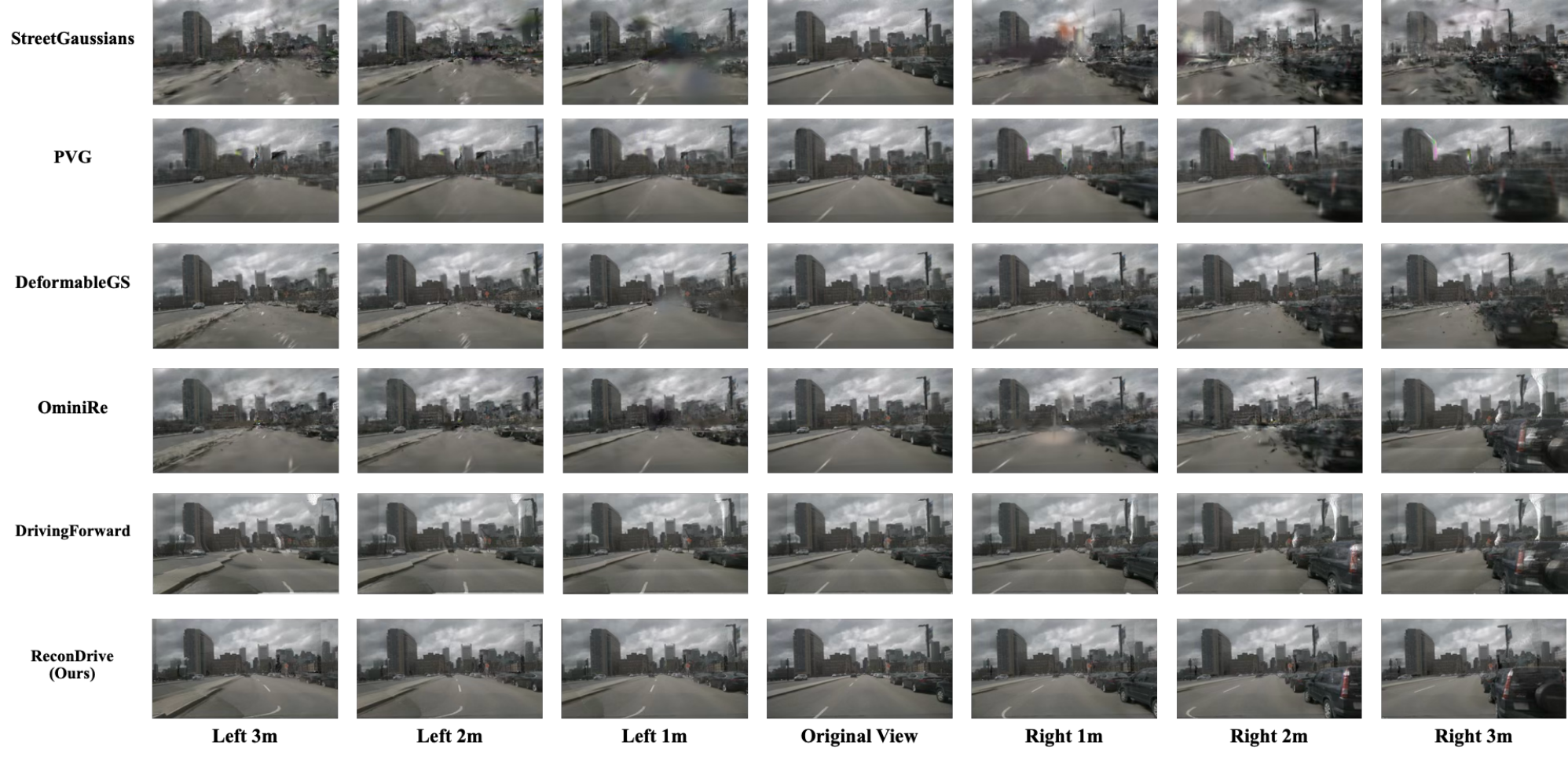}
	\caption{\textbf{Visual Comparisons of Scene Reconstruction and Novel-View Synthesis on Scene-0098 (from 14 Selected Validation Scenes).}
    Consistent with prior results, our ReconDrive maintains high-quality rendering in both scene reconstruction (Original View) and novel-view synthesis (1–3m lateral movement) compared to per-scene optimization methods (Street Gaussians, PVG, DeformableGS, OminiRe).}\label{fig:comparison-vis-3}
\end{figure*}

\end{document}